%% file: coling_latex.tex
\pdfoutput=1

\documentclass[11pt]{article}

\usepackage[preprint]{coling}

\usepackage{times}
\usepackage{latexsym}

\usepackage[T1]{fontenc}

\usepackage[utf8]{inputenc}

\usepackage{microtype}

\usepackage{inconsolata}

\usepackage{graphicx}

\usepackage{booktabs}
\usepackage{arydshln}
\usepackage{multirow}
\usepackage{amsmath}
\usepackage{cleveref}

\usepackage{graphicx}
\usepackage{caption}
\usepackage{subcaption}

\title{Enriching Datasets with Demographics through Large Language Models:\\\emph{What's in a Name?}}

\author{\textbf{Khaled AlNuaimi}$^*$$^1$$^,$$^2$, 
\textbf{Gautier Marti}$^*$$^2$, 
\textbf{Mathieu Ravaut}$^*$$^2$,  
\textbf{Abdulla AlKetbi}$^1$$^,$$^2$,\\
\textbf{Andreas Henschel}$^1$, 
\textbf{Raed Jaradat}$^1$\\
$^1$ Khalifa University, Abu Dhabi, UAE\\
$^2$ Abu Dhabi Investment Authority (ADIA), Abu Dhabi, UAE\\
}

\begin{document}

\maketitle

\def\thefootnote{*}\footnotetext{Authors contributed equally}\def\thefootnote{\arabic{footnote}}

\begin{abstract}

Enriching datasets with demographic information, such as gender, race, and age from names, is a critical task in fields like healthcare, public policy, and social sciences. Such demographic insights allow for more precise and effective engagement with target populations. Despite previous efforts employing hidden Markov models and recurrent neural networks to predict demographics from names, significant limitations persist: the lack of large-scale, well-curated, unbiased, publicly available datasets, and the lack of an approach robust across datasets. This scarcity has hindered the development of traditional supervised learning approaches. In this paper, we demonstrate that the zero-shot capabilities of Large Language Models (LLMs) can perform as well as, if not better than, bespoke models trained on specialized data. We apply these LLMs to a variety of datasets, including a real-life, unlabelled dataset of licensed financial professionals in Hong Kong, and critically assess the inherent demographic biases in these models. Our work not only advances the state-of-the-art in demographic enrichment but also opens avenues for future research in mitigating biases in LLMs\footnote{We will release all LLM-generated annotations}.


\end{abstract}

\input{Sections/1_introduction}

\input{Sections/2_related_work}

\input{Sections/3_setup}

\input{Sections/4_experiments}

\input{Sections/5_analysis}

\input{Sections/6_conclusion}

\section*{Limitations}

Our work, despite evaluating a large number of models, presents several limitations, some of which may be tackled in future work. 

First, we are limited by the quality of the data which we use. In the Wikipedia dataset, nationality annotations are automatically scrapped, therefore they are noisy. We partially clean them through prompting several LLMs to at least ensure that each entry corresponds to a human name. Better cleaning would be achieved by prompting LLMs with the entire Wikipedia page content. Besides, this Wikipedia dataset only contains pages in English. A more global dataset could be collected by also considering individuals without an English page but with pages in other languages, such as Chinese.

Next, is the ever-prevailing issue of data contamination in LLMs. Model behavior and performance might change if the LLM has been exposed to the data during pre-training. Wikipedia content is largely included in pre-training data dumps for most LLMs, and thus some content has been memorized, which undoubtedly has happened on some entries of the Wikipedia dataset which we used. While the other two datasets of Florida Voters and Hong Kong SFC datasets are unlikely to be contaminated due to more restricted access, there is still a non-zero chance of contamination. 

Lastly, we restrict all evaluations to the zero-shot setup. We expect significantly better performance when fine-tuning LLMs, especially for the task of birth year or age prediction. However, such an endeavour requires GPU resources out of our scope.

\section*{Acknowledgements}

We warmly thank Hien Ngo from ADIA for his help synchronizing inference with several LLMs.

\bibliography{custom}

\appendix

\input{Sections/a_parsing}

\end{document}

%% file: Sections/1_introduction.tex
\section{Introduction}

The rise of Large Language Models (LLMs) has marked a significant milestone in the evolution of artificial intelligence, particularly in natural language processing (NLP). Since the introduction of the Transformer architecture in 2017 \cite{vaswani2017attention}, LLMs have undergone rapid advancements, culminating in the development of models like GPT-3 \cite{brown2020language}, ChatGPT \cite{ouyang2022training} or Claude, which have demonstrated unprecedented capabilities in generating human-like text in zero-shot, bypassing the need for supervised tuning. These models have become ubiquitous in various applications, from chatbots to content creation tools, and are now essential in tasks such as summarizing lengthy documents \cite{goyal2022news,chang2023booookscore}, conducting information retrieval \cite{lewis2020retrieval}, assisting in code generation \cite{li2022competition}, and even solving complex mathematical problems \cite{trinh2024solving}.

Beyond their prowess in text generation, LLMs have ushered in a new paradigm in data generation \cite{schick2021generating,gupta2023targen}. The quality of LLM-generated content has reached a level where it can rival or even surpass human-generated data in certain contexts. For instance, instruction-tuning with LLM-curated or LLM-generated data points has been shown to improve performance on various NLP tasks, sometimes outperforming instruction-tuning with human-generated data \cite{xu2023wizardlm}. In specific tasks such as abstractive news summarization \cite{zhang2020pegasus}, human annotators have even rated LLM-generated labels as higher in quality than existing human labels \cite{zhang2024benchmarking}.

While LLMs have demonstrated impressive generation capabilities, their application to enriching datasets with demographic information—such as gender, race, and age—remains unexplored. Our study is the first to explore LLMs’ potential in enriching datasets with demographic information, addressing a critical gap in the field. This task is particularly important in areas where demographic data drives decision-making, such as healthcare, social sciences, and public policy. Demographics prediction presents unique challenges due to the vast cultural, linguistic, and regional variations in naming conventions. Moreover, the potential biases in LLMs \cite{bender2021dangers,kotek2023gender,ravaut2024much} could have far-reaching implications when applied to demographic data generation, affecting fairness and accuracy.

In this paper, we tackle demographics enhancement through zero-shot LLM prompting, using the individual's \emph{name} as only input variable. Our contributions are threefold: 

\begin{enumerate}

\item We demonstrate that modern \emph{zero-shot} LLMs outperform previous \emph{supervised} approaches, including hidden Markov models and recurrent neural networks, in generating demographic data from names.

\item We reveal critical biases in current LLMs, particularly their tendency to underestimate the age of individuals, often by more than a decade. This limitation has significant implications for age-sensitive applications, such as healthcare and marketing, where inaccurate age predictions can distort insights and lead to flawed decisions regarding treatment, resource allocation, and targeted campaigns.

\item We analyze, enrich, and release a novel dataset that focuses on the first and last names of finance professionals in Hong Kong, addressing the gap in non-Western demographic datasets, particularly those with a focus on Asian populations.

\end{enumerate}

These contributions not only pioneer the use of LLMs for demographics enrichment but also provide essential resources for future research, particularly in addressing biases and improving demographic predictions.

%% file: Sections/2_related_work.tex
\section{Related Work}

\subsection{Predicting Demographic Attributes from Names}
The task of predicting demographic attributes, such as race and ethnicity, from names has been a longstanding challenge, first explored in the early 1990s \cite{coldman1988classification,choi1993use,abrahamse1994surname}, primarily in fields like epidemiology and public policy. In recent years, this task has gained relevance in a broader range of domains, including social science research \cite{martiniello2022signaling} and machine learning \cite{wong2020machine,jain2022importance}.

Early methods for demographic prediction typically relied on static datasets, such as the U.S. Census Bureau's list of popular surnames\footnote{\url{https://www.census.gov/topics/population/genealogy/data.html}}, combined with basic statistical inference techniques. These methods, however, suffered from several key limitations. They were overly dependent on last names, which are heavily skewed towards non-Hispanic White populations, and the datasets themselves were updated infrequently, typically once every decade, making them slow to reflect important demographic shifts.

To address these limitations, more recent approaches have turned to supervised machine learning techniques. However, they remain heavily reliant on U.S.-centric datasets, and often fail to capture the cultural and linguistic diversity of naming conventions worldwide, limiting their generalizability to other regions.

\subsection{Existing Datasets and Their Limitations}

Existing datasets used for demographics prediction models, such as the U.S. Census Bureau’s list of popular last names and voter registration data, suffer from several limitations that hinder their generalizability. The U.S. Census data is skewed heavily towards non-Hispanic White individuals, with over 82\% of unique last names representing this demographic, and it excludes first names, which are crucial for more nuanced demographic distinctions. Additionally, voter registration data \citet{chintalapati2018predicting,parasurama2021racebert}, while more comprehensive in including both first and last names, is limited geographically, often lacks precise or consistent coding of race categories, and may not represent the entire population due to the voluntary nature of voter registration. Furthermore, Wikipedia-based datasets, though used in some studies to infer ethnicity from names \cite{ambekar2009name}, exhibit biases due to the over-representation of well-known individuals (75\% White, 80\% Male), making them less representative of the general population, and calling for the use of other, more diverse datasets.

\subsection{Machine Learning Approaches}

Recent advancements in demographics prediction have shifted from traditional models like Random Forests, Gradient Boosting, and k-NN \cite{chintalapati2018predicting}, which often relied on n-grams \cite{lee2017name}, to transformer-based models such as RaceBERT \cite{parasurama2021racebert}. RaceBERT, trained on U.S. voter registration data, outperforms earlier LSTM and RNN models in predicting race categories by better handling the nuances of both first and last names. While LSTMs demonstrated reasonable accuracy \cite{chintalapati2018predicting}, transformer models have shown superior generalization across diverse datasets.

\subsection{Novelty of Our Approach}

To the best of our knowledge, this study is the first to apply LLMs for demographic enrichment from names, addressing limitations in previous work that relied primarily on U.S. Census race categories and limited datasets. In contrast, we adopt a more global perspective by incorporating data from diverse, non-Western, contexts, particularly in Asia, and extend demographic inference beyond race and ethnicity to include variables such as gender or age.

Our approach also leverages a diverse range of LLMs, both open and closed sources, and from Western and Chinese providers. This dual focus on diverse data and models allows us to analyze regional and model-specific biases, providing a deeper understanding of the capabilities and limitations of LLMs in demographics prediction.






%% file: Sections/3_setup.tex
\section{Task}
\label{sec:task}

In this study, we perform model inference for diverse demographic variables using an individual's name as only relevant context. Formally, given an individual's name noted $X^{(i)}$, a model noted $f_{\theta}$, and a demographic variable $\mathbf{Y}$ taking discrete values, we prompt the model to predict the correct demographic value $Y^{(i)}$:

\begin{equation}
    f_{\theta}(\text{prompt}(X^{(i)})) = \hat{Y}^{(i)}
\end{equation}

We measure performance by comparing predicted and ground-truth class labels $\hat{Y}^{(i)}$ and $Y^{(i)}$, respectively. As examples, $X^{(i)}$ may take values \texttt{John Doe} ; $f_{\theta}$ may be GPT-4 ; and $\mathbf{Y}$ may represent \emph{Gender} and take values in the space $\mathbf{Y} \in \{\text{Male}, \text{Female}\}$.

We run all model inference with frozen, off-the-shelf LLMs, without using in-context-learning. We generate the response by sampling with a Temperature of 0 (mimicking greedy decoding). All required demographic variables are packed within the same prompt, and we guide the model by specifying the possible class values or expected format. For instance when predicting demographic variables $\{\emph{\text{Country of Origin}}, \emph{\text{Nationality}}, \emph{\text{Gender}}, \emph{\text{Race}}, \\
\emph{\text{Birth Date}}\}$, we use the following prompt:

\begin{verbatim}
f"""Given the full name of a person: 
{fullname}, please determine
the following details:
        
    1. The most likely country of origin, 
    represented by its ISO 3166-1 alpha-3 
    code (e.g., 'USA', 'GBR').
    2. The most likely nationality, also 
    represented by its ISO 3166-1 alpha-3 
    code.
    3. The gender of the person, reported 
    as 'M' for male or 'F' for female.
    4. The race of the person, choosing 
    from one of the following categories: 
    ['Hispanic', 'White, Not Hispanic', 
    'Black, Not Hispanic', 'Other', 
    'Asian Or Pacific Islander'].
    5. The estimated birth date, provided 
    in the format 'mm/dd/yyyy'.
    
Please return the information in the exact
format below:
    
    Country of Origin: [ISO3 code]
    Nationality: [ISO3 code]
    Gender: [M/F]
    Race: [Race Category]
    Birth Date: [mm/dd/yyyy]
    
Provide only the information requested, 
with no additional text or explanations."""
\end{verbatim}

\input{Tables/datasets}

%% file: Tables/datasets.tex
\begin{table*}[]
\centering
\resizebox{\textwidth}{!}{%
\begin{tabular}{llll}

\toprule 

\textbf{Name}               
& \textbf{Size} 
& \textbf{Split}                    
& \textbf{Annotations} \\

\midrule 

Florida Voters Registration 2022   & 15,009,273  & We subsample randomly 100k data points. & \{Gender, Birth Date, Race\} (all self-reported). \\
Wikipedia Persons                  & 1,112,905 & 890,249 / 111,287 / 111,369 existing train/dev/test split. & \{Nationality\} (automatically parsed). \\
Licensed Hong Kong SFC professionals   & 519,860 & We do not split it. There are 117,232 unique individuals. & None. \\

\bottomrule

\end{tabular}
}
\vspace{-0.5em}
\caption{High-level description of the datasets that we use.}
\label{tab:datasets}
\vspace{-1.0em}
\end{table*}

%% file: Sections/4_experiments.tex
\section{Experiments}

\subsection{Setup}

\begin{figure}[t]
    \centering
    \includegraphics[width=\columnwidth]{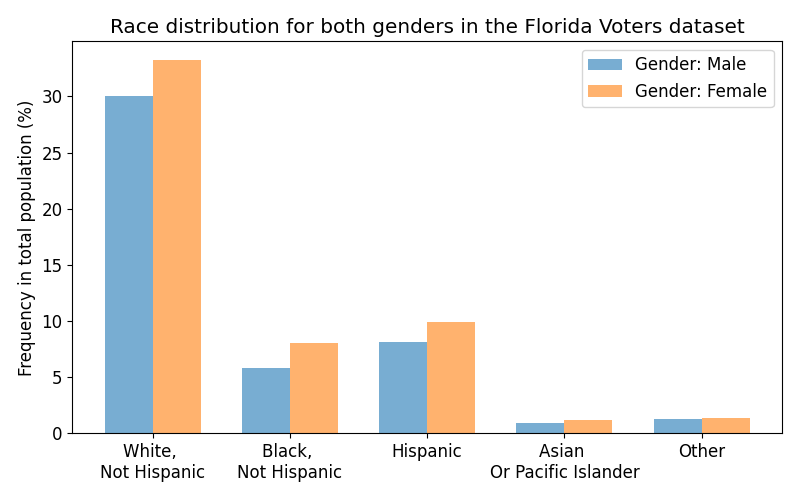}
    \vspace{-2.0em}
    \caption{Race distribution split by gender on the Florida Voters test set. Race is reduced from nine to five classes, as in prior work.}
    \label{fig:florida_demographics}
    \vspace{-1.5em}
\end{figure}

\begin{figure}[t]
    \centering
    \includegraphics[width=\columnwidth]{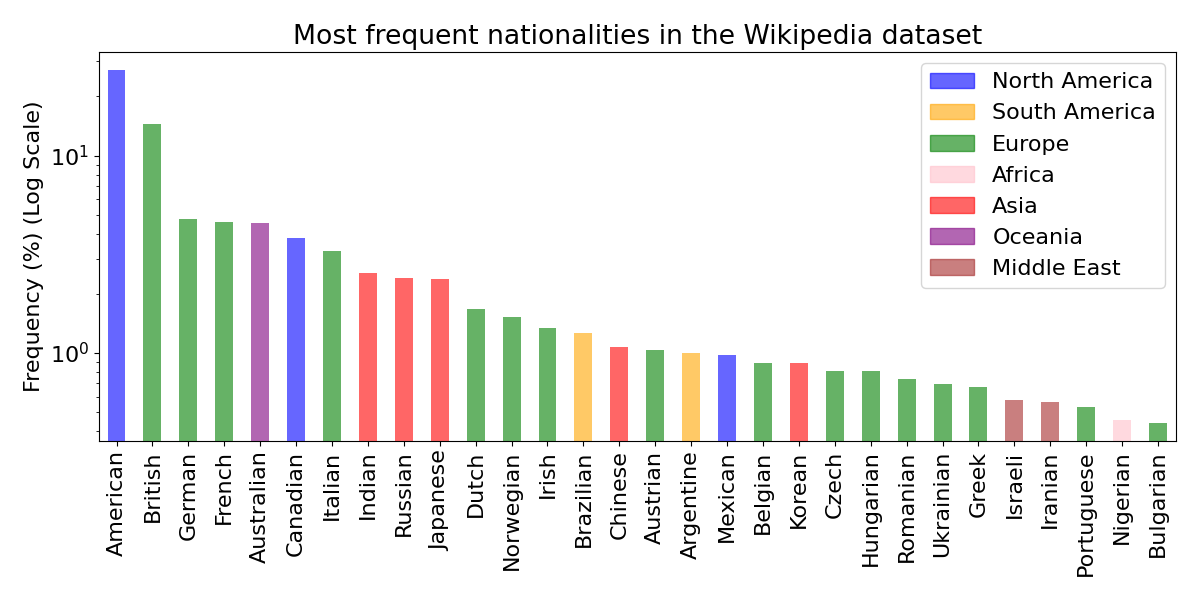}
    \vspace{-1.0em}
    \caption{Nationality distribution on the Wikipedia test set. The distribution is long-tail and skewed towards English-speaking countries and Europe. The top 30 nationalities displayed account for 87\% of data points.}
    \label{fig:wikipedia_top_nationalities}
    \vspace{-1.0em}
\end{figure}

\paragraph{Datasets}
We run inference on datasets with varying level of annotations. We first use the Florida Voters Registration 2022 dataset, following prior work \cite{chintalapati2018predicting,parasurama2021racebert}. This dataset contains self-reported Gender (two options: Male and Female), as well as self-reported Race, with nine options, and birth date. We subsample 100,000 data points randomly from the test set to run inference. The gender and race distribution for the Florida Voters dataset are shown in \Cref{fig:florida_demographics}.
Next, we use the dataset from Wikipedia with nationality annotations introduced by the \emph{name2nat} Python package \cite{park2018name2nat}. The most common nationalities are displayed in \Cref{fig:wikipedia_top_nationalities}.
Lastly, we also apply LLMs on a dataset containing information on finance professionals licensed by Hong Kong Securities \& Futures Commission (SFC)\footnote{\tiny \url{https://www.sfc.hk/en/Regulatory-functions/Intermediaries/Licensing/Register-of-licensed-persons-and-registered-institutions}}. 
A high-level description of all datasets is shown in \Cref{tab:datasets}.

\paragraph{Data cleaning}
Following prior work on the Florida Voters dataset \cite{chintalapati2018predicting}, we reduce Race options to five classes: \{White (Not Hispanic), Black (Not Hispanic), Hispanic, Asian or Pacific Islander, Other\}.
On the Wikipedia dataset, we noticed that the dataset contains other entries than people (horses, places, events around people's death), but also not legal birth names such as artists taken names, etc. To clean the dataset and only keep legal birth names, we run inference with four powerful LLMs - Claude-3-Haiku, Claude-3.5-Sonnet, GPT-3.5-turbo and GPT-4o - to predict whether each entry's name is a valid human name. Using a voting mechanism, with scores 0.15, 0.35, 0.20 and 0.30 for each model, respectively, we discard data points where the validity score is below 0.75, corresponding to 998 data points ($\leq 0.9$\% of the dataset).
We do not perform specific data cleaning on the Hong Kong SFC dataset.

\paragraph{LLMs}
We leverage a large variety of LLMs (12 in total), from both open-source and closed-source categories. 
On the open-source side, we use Mistral AI's Mistral-7B-Instruct (version 0.3) \cite{jiang2023mistral}, Alibaba's Qwen-2-7B-Instruct \cite{yang2024qwen2}, Meta's Llama-3-8B-Instruct and Llama-3.1-8B-Instruct \cite{dubey2024llama}, Yi.AI
s Yi-1.5-9B-Chat \cite{young2024yi}, and Google's Gemma-2-9B-it \cite{team2024gemma}. For all these open-source models, we download weights through HuggingFace transformers \cite{wolf2020transformers} and perform inference locally through vLLM \cite{kwon2023efficient} on 4 Nvidia A10G 24GB  cards.
On the closed-source side, we leverage Mistral AI's flagship Mistral-large model\footnote{\url{https://mistral.ai/news/mistral-large/}}, Cohere's Command\footnote{\url{https://cohere.com/command}}, Anthropic's Claude-3-Haiku and Claude-3.5-Sonnet\footnote{\url{https://www.anthropic.com/claude}}, and OpenAI's flagship models GPT-3.5-turbo and GPT-4o \cite{achiam2023gpt}. For these closed models, we access their respective paying API through LiteLLM\footnote{\url{https://github.com/BerriAI/litellm}}.
\Cref{tab:LLMs} summarizes the LLMs used in this paper, with publicly available information.

\input{Tables/llms}

\paragraph{Inference}
We prompt all LLMs with the same prompt template on each dataset, illustrated in \Cref{sec:task}.
We prompt for the variables with available labels \{gender, birth date, race\} on Florida Voters dataset. 
On the Wikipedia dataset, we conduct two types of inference: (1) a \emph{simple} inference that predicts \{nationality, gender\}, and (2) a \emph{complex} inference that predicts \{nationality, country of origin, race, gender, birth date\}.
On the Hong Kong SFC dataset, we ask LLMs to predict \{nationality, country of origin, ethnicity, gender, age\} ; and analyze agreement between LLMs due to the lack of annotations. 
We use minimal parsing on all datasets, just checking for the presence of expected fields in the LLM output string. As shown in \Cref{sec:appendix_parsing}, most LLM outputs are in the expected format, except on rare cases where we do not report performance.

\paragraph{Evaluation}
We evaluate gender, race and nationality prediction with \textbf{accuracy} ; and measure performance on age prediction by the \textbf{mean absolute error} (MAE) between predicted and ground-truth birth years. 
In annotated setups, we compute a \emph{Random} baseline consisting in shuffling ground-truth predictions. We also report a \emph{Most Frequent} baseline on classification use cases, and similarly an \emph{Average} baseline for the regression on birth date.

\subsection{Results}

\subsubsection{Gender prediction}

\Cref{tab:gender} presents LLM performance at predicting gender on the Florida Voters dataset. The binary-class dataset is relatively balanced (54\% self-reported Female, 46\% self-reported Male), and we notice very high accuracy for all 12 LLMs, ranging from 0.96 to 0.99 for Claude-3.5-Sonnet and GPT-4o. Besides, despite a slight decrease on Asian race, accuracy stays strong and above 0.85 across all races, for all LLMs. \textbf{We conclude that LLMs are mostly able to predict a person's gender solely based on the name}.

\subsubsection{Birth date prediction}
\label{subsubsec:birth_date}

\begin{figure}[t]
    \centering
    \includegraphics[width=\linewidth]{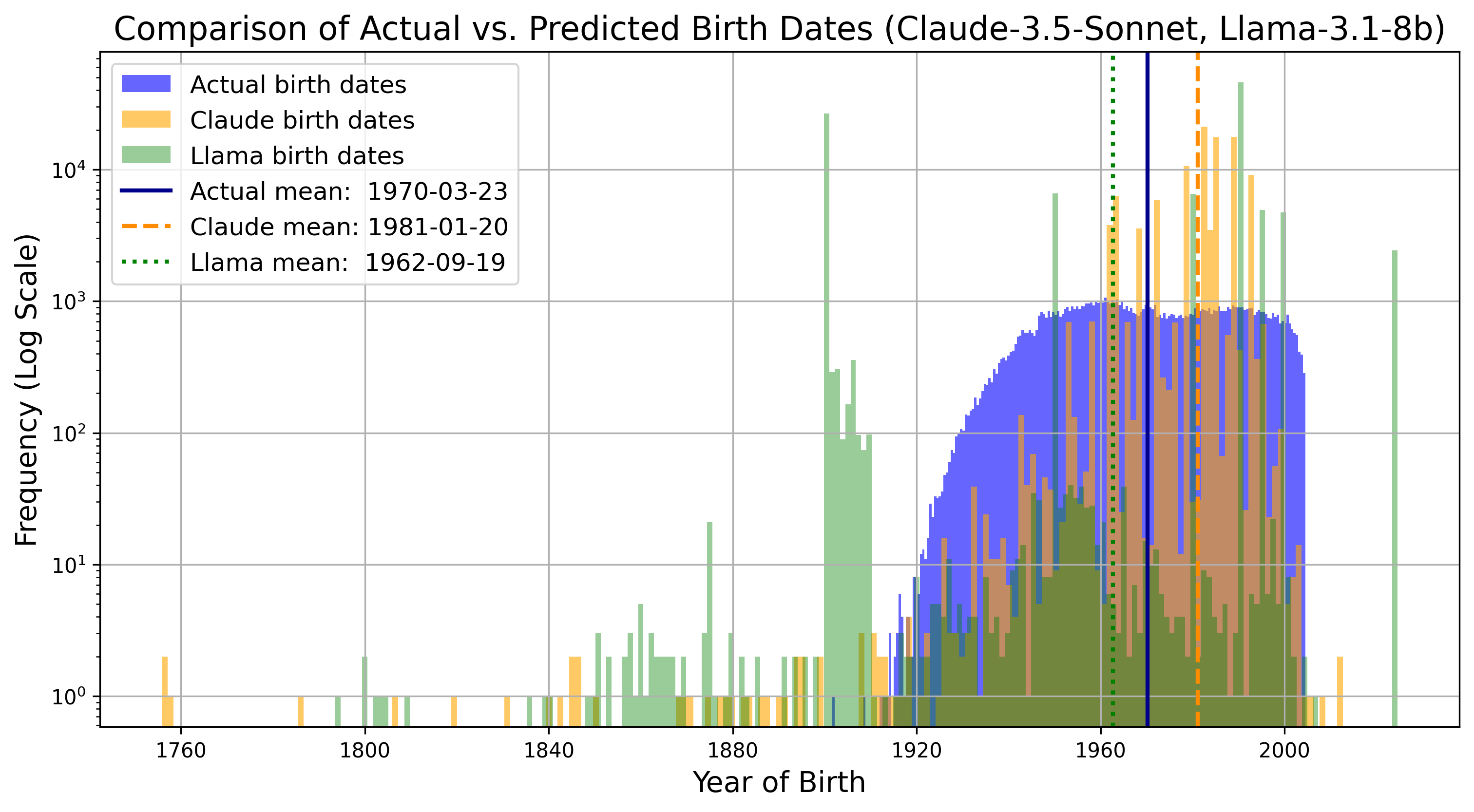}
    \caption{Comparison of actual vs. predicted \textbf{birth dates} (Claude-3.5-sonnet, Llama-3.1-8b) on Florida Voters.}
    \label{fig:birth_dates}
\end{figure}

We also use the available ground truth from the Florida Voters dataset and compute the mean absolute error between predicted and ground-truth birth years in \Cref{tab:birth_year}. LLMs perform poorly (especially open-source ones), and are not able to consistently improve on trivial baselines. We first illustrate the predicted distribution of one of the worst-performing (Llama-3.1-8B-Instruct) and the best-performing (Claude-3.5-Sonnet) LLMs in \Cref{fig:birth_dates}. Llama-3.1-8B-Instruct is completely off-range, but Claude-3.5-Sonnet better matches the ground-truth distribution shape. All LLMs generate historical dates prior to the nineteenth century, and a pronounced bias for a few round dates such as 1900 or 1990. They also predict more recent dates, except Gemma-2 predicting mostly 1900.
\textbf{We conclude that LLMs are not capable of predicting a birth date from a name. LLMs are biased towards round dates or more recent dates.}

\subsubsection{Race prediction}

In \Cref{tab:race}, we display LLM accuracy at predicting race on the Florida Voters dataset. Most LLMs show a zero-shot accuracy in the 0.75-0.85 range, on par with previously reported results with \emph{fine-tuned} machine learning models such as Random Forest or LSTM. While these fine-tuned models show very poor accuracy on under-represented groups Asian and Other, LLMs hold strong, with GPT-4o showing 0.74 accuracy on Asian race group. \textbf{We conclude that zero-shot LLMs outperform other fine-tuned supervised machine learning baselines at predicting racial group.}

\subsubsection{Nationality prediction}

Lastly, we present results on the more complicated task of predicting nationality based on name, with 166 classes present in the test set. Table~\ref{tab:nationality} presents overall nationality prediction accuracies on the Wikipedia dataset as well as breakdowns by gender for the simple inference, and by gender and race for the complex inference. On this task, accuracy is lower than on racial prediction. We notice clearly superior performance by closed-source LLMs, especially Claude and GPT series, with GPT-4o clearly stronger. Besides, \textbf{open-source LLMs benefit notably from the complex, multi-task inference setup}, gaining on average 15\% of accuracy. This finding echoes the line of research decomposing prompts in multiple steps, and enforcing self-consistency of zero-shot outputs \cite{wei2022chain,zhou2022prompt,wang2022self}. However, we notice that more powerful, closed-source models do not benefit from the multi-tasking of the complex inference setup. 

These results underscore the interplay between the biases inherent in LLMs and those present in the Wikipedia dataset. Due to Wikipedia being largely used in pre-training corpora \cite{raffel2020exploring,touvron2023llama}, many of the models may have been exposed to this data during pre-training, which could influence their performance in predicting certain nationalities. In the following, we delve into the existing biases of the Wikipeia Persons dataset which we used.

\input{Tables/florida_gender_prediction}

\input{Tables/florida_birth_date_prediction}

\input{Tables/florida_race_prediction}

\input{Tables/wikipedia_nationality_prediction}

\paragraph{Gender Bias} The dataset exhibits a significant gender imbalance, with only 20\% of individuals identified as female. This under-representation likely reflects broader societal biases, particularly within historical records and Wikipedia entries, where notable figures are predominantly male. Despite this disparity, LLMs demonstrated consistent performance in predicting nationality across both genders. However, it is crucial to recognize that these findings may not be globally representative. The dataset's focus on the Western, English-speaking world limits its generalizability, as naming conventions in other regions, such as China, include a higher prevalence of unisex names, which may present additional challenges for gender classification.

\paragraph{Race Breakdown:} The dataset exhibits a significant skew towards American entries, with 27\% of individuals identified as American (see Figure~\ref{fig:wikipedia_top_nationalities}). LLMs achieved high accuracy (74-82\%) when predicting the nationality of "Black" individuals, particularly those from the United States (93\%). This reflects a dataset bias, where approximately 44\% of the "Black" individuals in Wikipedia are listed as American. As a result, LLMs performed well in predicting the nationality of Black Americans.

Although "Black" individuals comprise only 5\% of the dataset, those classified as "White" represent around 75\%. Of this group, nearly half (49\%) are associated with either the United States or the United Kingdom, while the remaining 51\% predominantly hail from other European countries, Australia, and Canada. This distribution underscores the dataset’s bias toward English-speaking Western nations, particularly the United States.

LLMs encountered challenges in accurately classifying the nationality of individuals with Hispanic names, with accuracy dropping to 50\% or lower for all LLMs except GPT-4o (at 56\%). This difficulty arises from confusion between individuals from Latin American countries and those labeled as American. 
The presence of strong diasporic communities in the United States, combined with historical patterns of migration, complicates the task of nationality classification based on names alone.

%% file: Tables/llms.tex
\begin{table*}[h!]
\centering
\scriptsize 
\setlength{\tabcolsep}{3pt} 
\resizebox{0.9\textwidth}{!}{
\begin{tabular}{lllllll}

\toprule 

\textbf{LLM Name} & \textbf{Provider} & \textbf{Openness} & \textbf{Parameters} & \textbf{Context Length (tokens)} & \textbf{Pre-training Cut-off Date} & \textbf{Pre-training Tokens} \\ 

\midrule 

Mistral-7B-Instruct-v0.3    & Mistral AI & Open & 7B & 32K & Prior to December 2023 & Undisclosed  \\ 
Qwen-2-7B-Instruct          & Alibaba & Open & 7B & 128K & Prior to June 2024 & 7T \\  
LLaMA-3-8B-Instruct         & Meta & Open & 8B & 8K & March 2023 & 16.55T  \\ 
LLaMA-3.1-8B-Instruct       & Meta & Open & 8B & 128K & December 2023 & 16.55T  \\ 
Yi-1.5-9B-Chat              & Yi.AI & Open & 9B & 4K & December 2023 & 3.1T  \\ 
Gemma-2-9B-it               & Google & Open & 9B & 8K & Prior to June 2024 & 8T  \\ 

\hdashline 

Mistral-Large      & Mistral AI & Closed & Undisclosed & 32K & Prior to December 2023 & Undisclosed  \\ 
Cohere-Command     & Cohere & Closed & Undisclosed & 4K & Prior to September 2023 & Undisclosed  \\ 
Claude-3-Haiku     & Anthropic & Closed & Undisclosed & 200K & August 2023 & Undisclosed   \\ 
Claude-3.5-Sonnet  & Anthropic & Closed & Undisclosed & 200K & August 2023 & Undisclosed \\ 
GPT-3.5-turbo      & OpenAI & Closed & Undisclosed & 16K & September 2021 & Undisclosed  \\ 
GPT-4o             & OpenAI & Closed & Undisclosed & 4K & October 2023 & Undisclosed \\ 

\bottomrule

\end{tabular}
}
\caption{Description of the LLMs utilized in our study, including their name, source type, model size, context length, pre-training cut-off date, and available details about pre-training data. The dash line separates open-source LLMs from closed-source ones.}
\label{tab:LLMs}
\end{table*}

%% file: Tables/florida_gender_prediction.tex
\begin{table}[t]
\centering
\scriptsize 
\setlength{\tabcolsep}{2pt} 
\resizebox{\columnwidth}{!}{%
\begin{tabular}{lcccccc}

\toprule

\textbf{Model} & \textbf{Overall} & \textbf{White} & \textbf{Black} & \textbf{Hispanic} & \textbf{Asian} & \textbf{Other} \\

\midrule        

\emph{Random}          & \emph{0.50} & \emph{0.50} & \emph{0.50} & \emph{0.50} & \emph{0.50} & \emph{0.50} \\
\emph{Most Frequent (Female)} & \emph{0.54} & \emph{0.53} & \emph{0.58} & \emph{0.55} & \emph{0.57} & \emph{0.53} \\ 

\midrule

Mistral-7B-Instruct    & \underline{0.98} & \underline{0.98} & 0.95 & \underline{0.98} & 0.92 & 0.96 \\ 
Qwen-2-7B-Instruct     & \underline{0.98} & \underline{0.98} & 0.95 & \underline{0.98} & 0.92 & 0.96 \\ 
Llama-3-8B-Instruct    & \underline{0.98} & \textbf{0.99} & \underline{0.96} & \underline{0.98} & 0.92 & 0.96 \\ 
Llama-3.1-8B-Instruct  & \underline{0.98} & \textbf{0.99} & \underline{0.96} & \underline{0.98} & 0.92 & 0.96 \\ 
Yi-1.5-9B-Chat         & 0.97 & \underline{0.98} & 0.94 & 0.97 & 0.90 & 0.95 \\ 
Gemma-2-9B-it          & \underline{0.98} & \textbf{0.99} & \underline{0.96} & \underline{0.98} & 0.93 & 0.96 \\

\hdashline 

Mistral-large          & 0.97 & \underline{0.98} & 0.91 & 0.97 & 0.87 & 0.94 \\
Cohere-Command         & 0.96 & 0.97 & 0.93 & 0.97 & 0.86 & 0.94 \\
Claude-3-Haiku         & \underline{0.98} & \textbf{0.99} & \underline{0.96} & \underline{0.98} & 0.93 & \underline{0.97} \\ 
Claude-3.5-Sonnet      & \textbf{0.99} & \textbf{0.99} & \textbf{0.97} & \textbf{0.99} & \textbf{0.95} & \textbf{0.98} \\ 
GPT-3.5-turbo          & \underline{0.98} & \textbf{0.99} & \underline{0.96} & \textbf{0.99} & 0.93 & \underline{0.97} \\
GPT-4o                 & \textbf{0.99} & \textbf{0.99} & \textbf{0.97} & \textbf{0.99} & \underline{0.94} & \textbf{0.98} \\ 

\bottomrule

\end{tabular}
}
\caption{Model accuracy at predicting \textbf{gender} (2 classes) on the Florida Voters dataset, split per racial group. Best numbers in bold, second best underlined.}
\label{tab:gender}
\vspace{-0.5em}
\end{table}

%% file: Tables/florida_birth_date_prediction.tex
\begin{table}[t]
\centering
\scriptsize 
\setlength{\tabcolsep}{2pt} 
\resizebox{\columnwidth}{!}{%
\begin{tabular}{lcccccc}

\toprule

\textbf{Model} & \textbf{Overall} & \textbf{White} & \textbf{Black} & \textbf{Hispanic} & \textbf{Asian} & \textbf{Other} \\

\midrule        

\emph{Random}          & \emph{21.9} & \emph{22.1} & \emph{21.7} & \emph{21.7} & \emph{21.0} & \emph{22.4} \\
\emph{Average year (1970)}    & \emph{16.2} & \emph{16.5} & \emph{15.6} & \emph{15.9} & \emph{15.2} & \emph{17.1} \\ 
\emph{Average year per Race}  & \emph{15.8} & \emph{16.1} & \emph{14.9} & \emph{15.3} & \emph{14.8} & \emph{16.2} \\ 

\midrule

Mistral-7B-Instruct    & - & - & - & - & - & - \\ 
Qwen-2-7B-Instruct     & 17.8 (+12.5) & 18.4 & 15.4 & 17.6 & 17.8 & 16.1 \\ 
Llama-3-8B-Instruct    & - & - & - & - & - & - \\ 
Llama-3.1-8B-Instruct  & 29.7 (-7.2) & 31.6 & 27.5 & 25.4 & 26.9 & 26.3 \\ 
Yi-1.5-9B-Chat         & \underline{16.5} (+7.6) & \underline{16.5} & 15.3 & 17.1 & 19.6 & 16.9 \\ 
Gemma-2-9B-it          & 69.9 (-69.4) & 67.4 & 74.2 & 74.4 & 73.7 & 74.7 \\ 

\hdashline 

Mistral-large          & - & - & - & - & - & - \\
Cohere-Command         & 19.9 (+16.5) & 21.2 & 17.0 & 18.0 & 20.1 & 18.2 \\
Claude-3-Haiku         & 16.6 (+10.9) & 17.7 & \underline{13.6} & \textbf{15.3} & \textbf{15.5} & \underline{15.1} \\
Claude-3.5-Sonnet      & \textbf{15.0} (+10.7) & \textbf{15.4} & \textbf{12.3} & \underline{15.8} & \underline{16.0} & \textbf{14.0} \\
GPT-3.5-turbo          & 19.6 (+16.4) & 21.5 & 15.8 & 16.8 & 17.4 & 16.7 \\
GPT-4o                 & 16.6 (+12.1) & 17.4 & 13.9 & 16.4 & 16.8 & 15.1 \\

\bottomrule

\end{tabular}
}
\caption{MAE at predicting \textbf{birth year} on the Florida Voters dataset, split per racial group. The "-" symbol indicates that the LLM generated a valid date format on too few data points. We also show in parenthesis the difference in years between average ground truth and average predicted year.}
\label{tab:birth_year}
\vspace{-0.5em}
\end{table}

%% file: Tables/florida_race_prediction.tex
\begin{table}[t]
\centering
\scriptsize 
\setlength{\tabcolsep}{2pt} 
\resizebox{\columnwidth}{!}{%
\begin{tabular}{lcccccc}

\toprule

\textbf{Model} & \textbf{Overall} & \textbf{White} & \textbf{Black} & \textbf{Hispanic} & \textbf{Asian} & \textbf{Other} \\

\midrule 

\emph{Random}           & \emph{0.45} & \emph{0.63} & \emph{0.14} & \emph{0.18} & \emph{0.02} & \emph{0.03} \\ 
\emph{Most Frequent (NH White)} & \emph{0.63} & \emph{1.00} & \emph{0.00} & \emph{0.00} & \emph{0.00} & \emph{0.00} \\
Random Forest*          & 0.72 & 0.94 & 0.25 & 0.66 & 0.17 & 0.02 \\
Gradient Boosting*      & 0.65 & \textbf{0.98} & 0.01 & 0.37 & 0.04 & 0.00 \\
LSTM*                   & 0.81 & 0.90 & \textbf{0.68} & 0.83 & 0.56 & 0.07 \\ 
Transformer*            & 0.66 & 0.94 & 0.03 & 0.50 & 0.08 & 0.00 \\  

\midrule 

Mistral-7B-Instruct    & 0.78 & 0.91 & 0.29 & 0.86 & 0.65 & 0.00 \\ 
Qwen-2-7B-Instruct     & 0.78 & \underline{0.95} & 0.13 & 0.85 & 0.42 & 0.00 \\ 
Llama-3-8B-Instruct    & 0.78 & \underline{0.95} & 0.06 & 0.85 & 0.36 & \underline{0.19} \\ 
Llama-3.1-8B-Instruct  & 0.79 & \underline{0.95} & 0.11 & 0.87 & 0.48 & 0.10 \\ 
Yi-1.5-9B-Chat         & 0.55 & 0.57 & 0.10 & 0.82 & 0.54 & \textbf{0.43} \\ 
Gemma-2-9B-it          & 0.79 & 0.94 & 0.14 & 0.88 & 0.55 & 0.12 \\

\hdashline 

Mistral-large          & 0.80 & 0.91 & 0.35 & 0.88 & 0.64 & 0.06 \\
Cohere-Command         & 0.71 & 0.79 & 0.25 & 0.90 & 0.15 & 0.10 \\
Claude-3-Haiku         & 0.80 & 0.93 & 0.22 & \underline{0.91} & 0.62 & 0.10 \\ 
Claude-3.5-Sonnet      & \underline{0.83} & 0.92 & 0.49 & \textbf{0.92} & \underline{0.72} & 0.06 \\ 
GPT-3.5-turbo          & 0.82 & 0.90 & 0.52 & 0.90 & 0.48 & 0.16 \\
GPT-4o                 & \textbf{0.84} & 0.92 & \underline{0.55} & 0.90 & \textbf{0.74} & 0.06 \\ 

\bottomrule

\end{tabular}
}
\caption{Model accuracy at predicting \textbf{racial group} (5 classes) on the Florida Voters dataset. *Baseline model results are taken from reported results in \cite{chintalapati2018predicting}.}
\label{tab:race}
\vspace{-0.5em}
\end{table}

%% file: Tables/wikipedia_nationality_prediction.tex
\begin{table*}[h]
\resizebox{\textwidth}{!}{
\begin{tabular}{l|ccc|cccccccc}

\toprule 

\multirow{2}{*}{\textbf{Model}} & \multicolumn{3}{c}{\textit{\textbf{Simple inference}}} & \multicolumn{8}{c}{\textit{\textbf{Complex inference}}}                                                                                            \\

                                & \textbf{Overall}   & \textbf{Male*}  & \textbf{Female*}  & \textbf{Overall} & \textbf{Male} & \textbf{Female} & \textbf{White} & \textbf{Black} & \textbf{Hispanic} & \textbf{Asian or P.I.} & \textbf{Other} \\

\midrule 

\emph{Random}           & \emph{0.11} & \_ & \_ & \emph{0.11} & \_ & \_ & \_ & \_ & \_ & \_ & \_ \\ 
\emph{Most Frequent (USA)} & \emph{0.27} & \_ & \_ & \emph{0.27} & \_ & \_ & \_ & \_ & \_ & \_ & \_ \\ 

\midrule 

Mistral-7B-Instruct     & 0.34 & 0.38 & 0.32 & 0.62 & 0.62 & 0.64 & 0.63 & 0.66 & 0.37 & 0.81 & 0.50 \\
Qwen-2-7B-Instruct      & 0.57 & 0.59 & 0.58 & 0.60 & 0.60 & 0.61 & 0.60 & 0.47 & 0.31 & 0.81 & 0.49 \\
Llama-3-8B-Instruct     & 0.57 & 0.59 & 0.58 & 0.67 & 0.67 & 0.68 & 0.69 & 0.76 & 0.40 & \underline{0.87} & 0.64 \\
Llama-3.1-8B-Instruct   & 0.60 & 0.62 & 0.60 & 0.69 & 0.69 & 0.69 & 0.71 & \underline{0.78} & 0.40 & \textbf{0.88} & 0.63 \\
Yi-1.5-9B-Chat          & 0.25 & 0.35 & 0.32 & 0.54 & 0.55 & 0.58 & 0.56 & 0.66 & 0.18 & 0.83 & 0.52 \\
Gemma-2-9B-it           & 0.67 & 0.67 & 0.67 & 0.65 & 0.65 & 0.66 & 0.67 & 0.74 & 0.37 & 0.78 & 0.46 \\

\hdashline 

Mistral-large           & 0.58 & 0.64 & 0.64 & 0.69 & 0.69 & 0.71 & 0.69 & 0.77 & 0.45 & 0.85 & 0.60 \\
Cohere-Command          & 0.42 & 0.55 & 0.55 & 0.49 & 0.50 & 0.47 & 0.47 & 0.49 & 0.22 & 0.70 & 0.57 \\
Claude-3-Haiku          & 0.64 & 0.65 & 0.62 & 0.67 & 0.67 & 0.66 & 0.68 & 0.74 & 0.40 & 0.81 & 0.54 \\
Claude-3.5-Sonnet       & \underline{0.70} & \underline{0.70} & \underline{0.70} & 0.70 & 0.69 & 0.71 & 0.69 & 0.80 & 0.47 & 0.83 & \underline{0.69} \\
GPT-3.5-turbo           & 0.69 & 0.69 & \underline{0.70} & \underline{0.72} & \underline{0.72} & \underline{0.74} & \underline{0.72} & 0.74 & \underline{0.49} & 0.86 & 0.65 \\
GPT-4o                  & \textbf{0.76} & \textbf{0.76} & \textbf{0.78} & \textbf{0.75} & \textbf{0.75} & \textbf{0.77} & \textbf{0.75} & \textbf{0.82} & \textbf{0.56} & 0.85 & \textbf{0.74} \\

\bottomrule   
\end{tabular}
}
\vspace{-0.5em}
\caption{Model accuracy at predicting the correct \textbf{nationality} on the Wikipedia test set (166 classes). We compare two setups: \emph{simple inference} in which we prompt the LLM to generate \{gender, nationality\} ; and \emph{complex inference} where the LLM has to generate \{gender, race, birth date, country of origin, nationality\}. Accuracy is shown on the whole set and by splitting across diverse demographics (predicted gender, predicted race). *Overall accuracy may not correspond to an average of accuracy split over demographics, as in some cases the model fails to generate a valid demographic field and we discard such data points.}
\label{tab:nationality}
\end{table*}

%% file: Sections/5_analysis.tex
\section{Analysis}

\subsection{Bias in Birth Year and Age}

In \Cref{subsubsec:birth_date}, we noticed mode collapse from most LLMs, which frequently predicted a "round" year of birth such as 1900 or 1990 on Florida Voters. LLMs were also skewed towards more recent dates. We now investigate if such pattern persist when predicting \emph{Age} instead of birth date. 

\begin{figure}[t]
    \centering
    \includegraphics[width=\columnwidth]{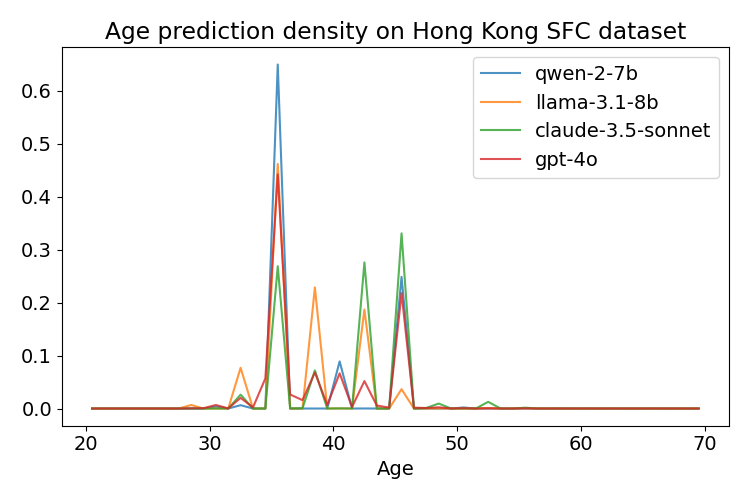}
    \vspace{-2.0em}
    \caption{Density of age prediction on the Hong Kong SFC professionals dataset, for four LLMs.}
    \vspace{-1.0em}
    \label{fig:hk_age}
\end{figure}

As shown in \Cref{fig:hk_age} for the Hong Kong SFC dataset, LLMs also present mode collapse when directly predicting the age. Indeed, LLMs mostly predict round ages such as 35 or 45 years old. For instance, Qwen-2-7B predicts 35 years old for more than 60\% of data points. Interestingly, this behavior also affects powerful LLMs like Claude-3.5-Sonnet and GPT-4o. We conclude that \textbf{predicting the birth date or age is very challenging for LLMs, and they will fall back to mode collapsing on a small set of round values for this task.}

\input{Tables/ensemble}

\begin{figure*}
    \centering
    \begin{subfigure}[b]{0.24\textwidth}  
        \centering
        \includegraphics[width=\linewidth]{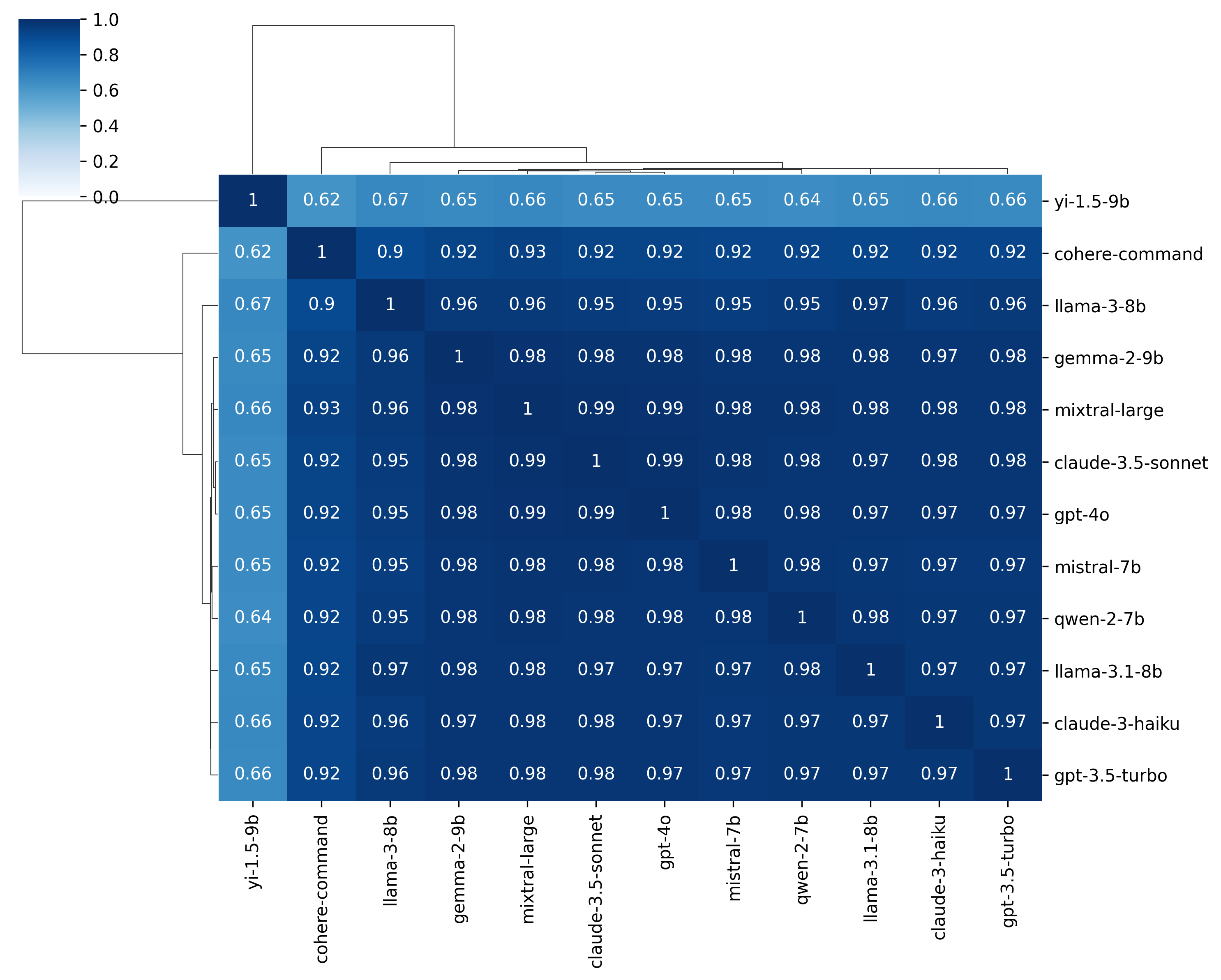}
        \caption{\tiny Race Agreement on Florida Voters}
        \label{fig:clustermap_race_predictions}
    \end{subfigure}
    \hfill
    \begin{subfigure}[b]{0.24\textwidth}  
        \centering
        \includegraphics[width=\linewidth]{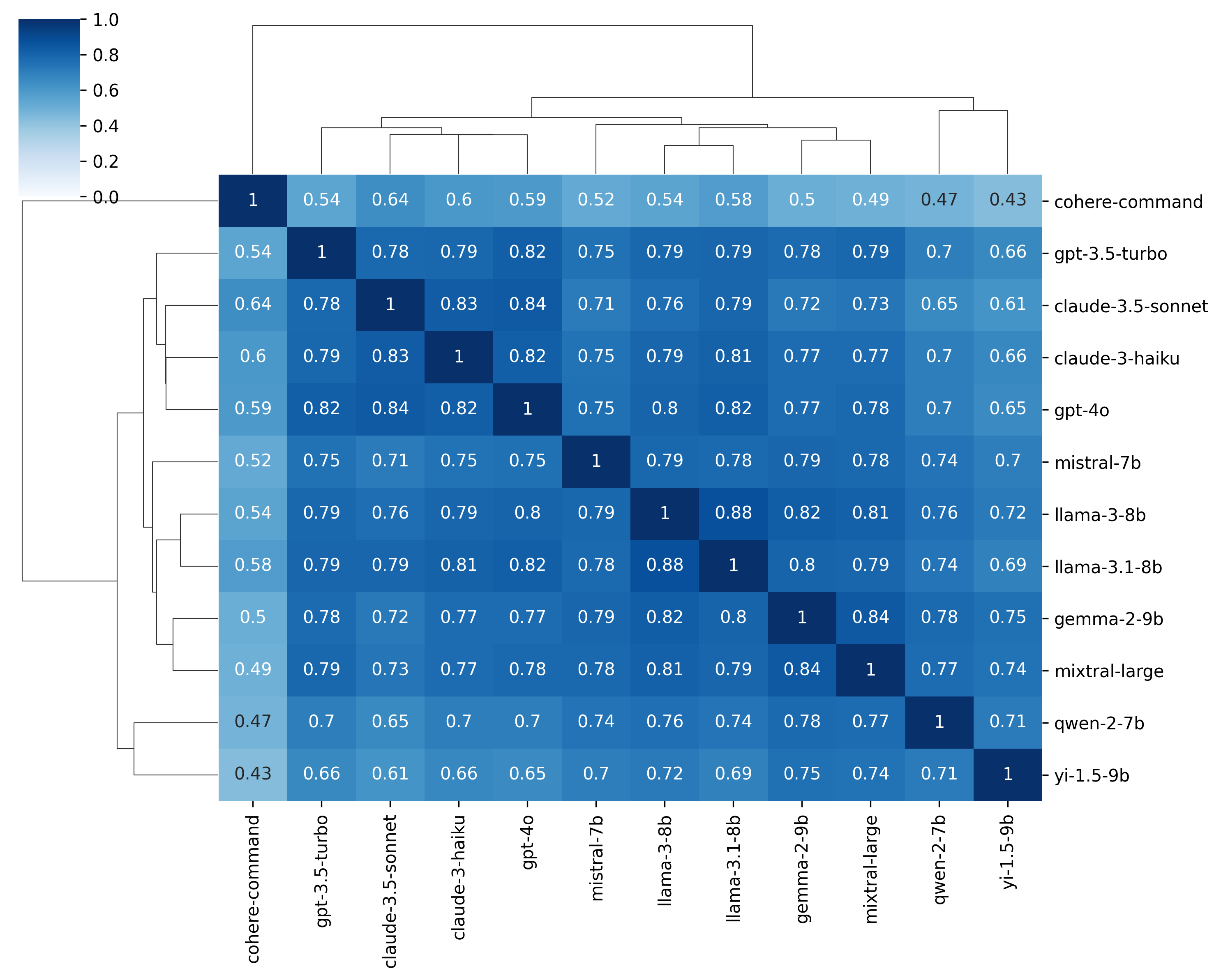}
        \caption{\tiny Nationality Agreement on Wikipedia}
        \label{fig:clustermap_nationality_predictions}
    \end{subfigure}
    \hfill
    \begin{subfigure}[b]{0.24\textwidth}  
        \centering
        \includegraphics[width=\linewidth]{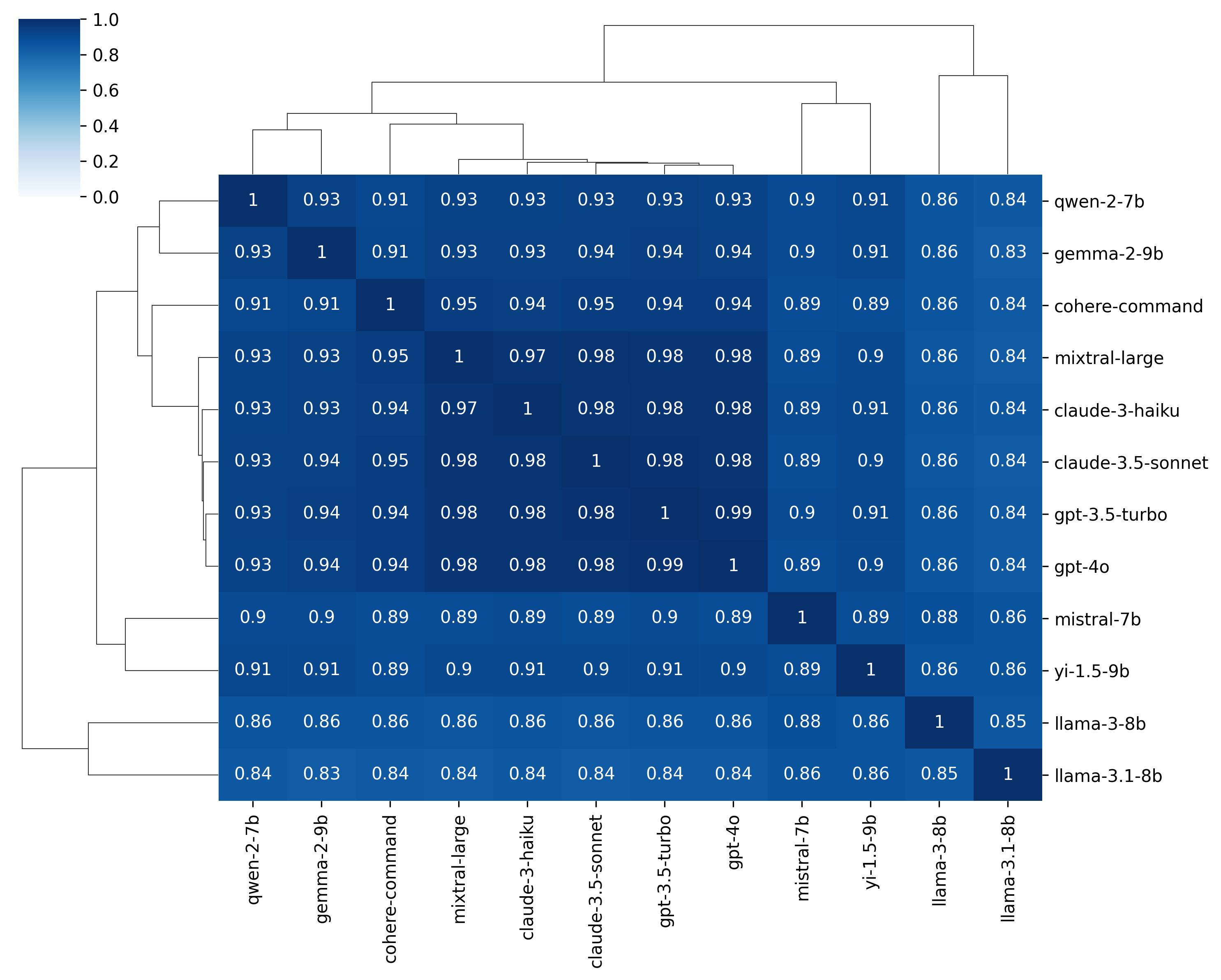}
        \caption{\tiny Ethnicity Agreement on HK SFC}
        \label{fig:clustermap_ethnicity}
    \end{subfigure}
    \hfill
    \begin{subfigure}[b]{0.24\textwidth}  
        \centering
        \includegraphics[width=\linewidth]{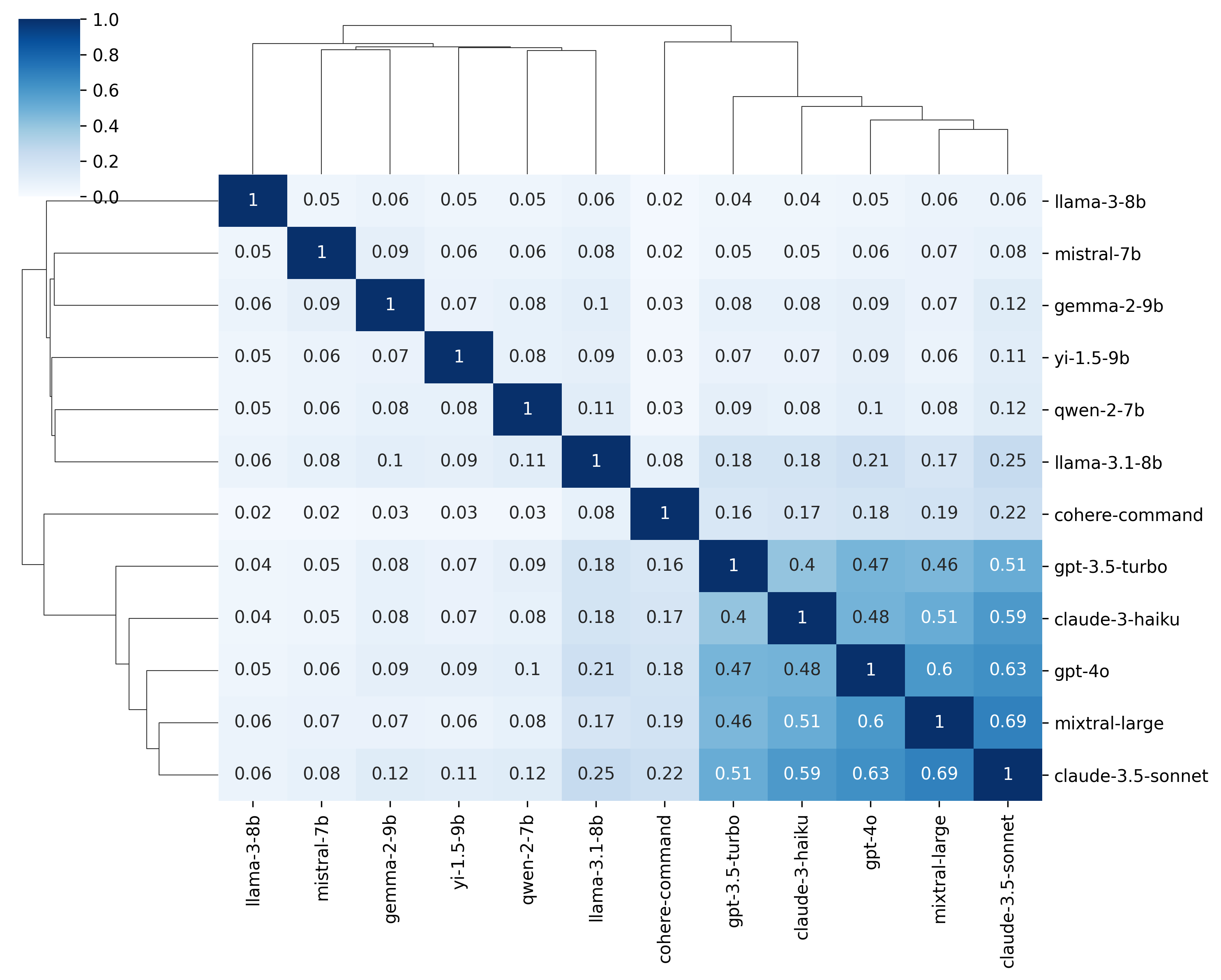}
        \caption{\tiny Age Agreement on HK SFC}
        \label{fig:llm_agreement_predicted_age}
    \end{subfigure}

    \caption{Hierarchical clustering of LLMs based on their agreement on predictions for the three datasets: Florida, Wikipedia, and HK SFC. Left to right: (a) Race, (b) Nationality (complex setup), (c) Ethnicity, and (d) Predicted Age agreement.}
    \vspace{-0.5em}
    \label{fig:combined_clustermaps}
\end{figure*}

\subsection{LLMs Agreement}

We analyzed the agreement between 12 LLMs by comparing their predictions. For classification tasks like gender, we used pairwise agreement to assess similarity, while for continuous predictions like age, we applied correlation. For ethnicity, where models generated a large number of unique ($\approx 1600$) but often similar outputs, we accounted for the non-orthogonal nature of classes by embedding the predictions using OpenAI's text-embedding-ada-002\footnote{\url{https://platform.openai.com/docs/guides/embeddings/embedding-models}} and calculating cosine similarity. Strong agreement was found for simpler tasks (gender and "fuzzy" ethnicity), with lower agreement for nationality and age. LLMs cluster by source type—open-source vs. closed-source—with a high-agreement cluster of Claude 3.5 Sonnet, GPT-4, GPT-3.5 Turbo, and Claude 3 Haiku. Mistral-large correlates moderately with this cluster. Pairwise agreement results are shown in Figure~\ref{fig:combined_clustermaps}.

We experimented with ensembling LLMs using majority voting, selecting the most frequent prediction or randomly choosing among ties. This was applied to supervised classification tasks using all 12 LLMs first, and then the top 3 performers. As shown in \Cref{tab:ensemble}, majority voting yields no performance improvement, which we attribute to the high correlation found above. This finding highlights the challenge of ensembling LLM outputs.

\subsection{Error Analysis}

\paragraph{Noisy Nationality Labels}
On Wikipedia nationalities, we previously noticed significantly lower accuracy on Hispanic names. We noticed that the ground truth dataset itself frequently misclassified Latino individuals as American, leading to discrepancies between labels and model predictions. In some instances, LLMs even correctly identified the nationality of these individuals, while the ground truth labels were in fact incorrect. For example, notable athletes such as Jailma de Lima (Brazilian track and field hurdler), Aixa Middleton González (Panamanian track and field athlete), Dania Pérez (Cuban cyclist), and Horacio Esteves (Venezuelan sprinter) were all misclassified as American in the dataset, despite the LLMs accurately predicting their Latin American nationalities.

\paragraph{Nationality vs Country of Origin}
Due to historical immigration patterns, many individuals from the United States, Canada, and Australia possess European surnames. However, their distinctive first names often enable LLMs to correctly infer their nationality, distinguishing it from the European origins suggested by their surnames. In rare cases, LLMs predict a nationality that diverges from the individual’s country of origin—this occurs in 0.8\% of instances. For example, Wolfgang K. H. Panofsky, born in Berlin, Germany, on April 24, 1919, became a U.S. citizen in 1942, and his nationality was correctly predicted as American. Similarly, in the case of Sho Yano, the model accurately predicted both his nationality as American and his exact birth date (October 22, 1990), despite his Japanese origins.
These instances suggest that LLMs may have memorized certain well-known individuals during pre-training. However, such memorization appears to be limited, as the distinction between predicted nationality and country of origin was observed in only a small portion (0.8\%) of the dataset.

%% file: Tables/ensemble.tex
\begin{table}[]
\resizebox{\columnwidth}{!}{%
\begin{tabular}{lcccc}

\toprule 

\textbf{Model} & \textbf{Gender} & \textbf{Race} & \textbf{Nat. (Simple)} & \textbf{Nat. (Complex)} \\

\midrule 

\emph{Random}              & \emph{0.50} & \emph{0.45} & \emph{0.11} & \emph{0.11} \\
\emph{Most Frequent}       & \emph{0.54} & \emph{0.63} & \emph{0.27} & \emph{0.27} \\
Best LLM                   & \textbf{0.99} & \textbf{0.83} & \textbf{0.76} & \textbf{0.75} \\
LLM ensemble (12 models)   & 0.98 & 0.80 & 0.72 & 0.72 \\
LLM ensemble (3 models)    & 0.98 & \textbf{0.83} & 0.75 & \textbf{0.75} \\

\bottomrule 

\end{tabular}
}
\caption{Accuracy of majority vote from a pool of LLMs on the classification tasks, compared to baselines and the best LLM for each task. \textbf{Nat.} is short for Nationality.}
\vspace{-0.5em}
\label{tab:ensemble}
\end{table}

%% file: Sections/6_conclusion.tex
\section{Conclusion}

In this paper, we demonstrated that LLMs are capable of accurately predicting the gender, race, or even nationality of a person, solely based on their name. They outperform previously reported supervised models and are more consistent across diverse population groups. In particular, Claude-3.5-Sonnet and GPT-4o exhibit the strongest performance in zero-shot demographic enrichment.

However, the task of predicting age or birth date remains more challenging. While there is evidence that certain trends in first names can offer clues for estimating the date of birth, current LLMs have not yet fully captured these patterns. LLMs are notably biased towards more recent birth dates and younger ages. This limitation suggests that further advancements in model training may be required for LLMs to better utilize such subtle correlations.

LLMs usher a new era of large-scale demographics annotation generation, which could significantly streamline many population-level interventions, such as in medicine. Moreover, these models could enhance transparency and accountability by identifying biases in media coverage and sentiment toward specific demographic groups in public discourse.



%% file: Sections/a_parsing.tex
\section{Parsing}
\label{sec:appendix_parsing}

\input{Tables/parsing}

In \Cref{tab:parsing}, we report the fraction of data points on which LLMs generate an output in the expected format for each prompted field, across all datasets and tasks. In the majority of cases, LLMs outputs are in the correct format, except for Mistral-7B-Instruct, Llama-3-8B-Instruct and Mistral-large, which struggle on some tasks, notably regarding birth date or age prediction.

%% file: Tables/parsing.tex
\begin{table*}[t]
\resizebox{\textwidth}{!}{%
\begin{tabular}{lccc|cc|ccccc|ccccc}

\toprule 

\textbf{}           & \multicolumn{3}{c}{\textbf{Florida Voters}}           & \multicolumn{2}{c}{\textbf{Wikipedia - simple}} & \multicolumn{5}{c}{\textbf{Wikipedia - complex}}                                                          & \multicolumn{5}{c}{\textbf{Hong Kong SFC}}                                                              \\

\textbf{LLM}        
& \textbf{Gender} & \textbf{Birth date} & \textbf{Race} 
& \textbf{Nationality} & \textbf{Gender}     
& \textbf{Nationality} & \textbf{Origin*} & \textbf{Race} & \textbf{Gender} & \textbf{Birth date} 
& \textbf{Nationality} & \textbf{Origin*} & \textbf{Ethnicity} & \textbf{Gender} & \textbf{Age} \\

\midrule 

Mistral-7B-Instruct   & 1.00 & \textcolor{red}{0.01} & 0.99 & 0.52 & 0.89 & 0.92 & 0.90 & 0.86 & 0.96 & \textcolor{red}{0.07} & \textcolor{red}{0.03} & \textcolor{red}{0.03} & 1.00 & 0.42 & 0.41 \\
Qwen-2-7B-Instruct    & 1.00 & 1.00 & 1.00 & 0.89 & 0.97 & 0.96 & 0.96 & 1.00 & 1.00 & 0.91 & 1.00 & 0.99 & 1.00 & 1.00 & 1.00 \\
Llama-3-8B-Instruct   & 1.00 & \textcolor{red}{0.00} & 1.00 & 0.86 & 0.98 & 0.98 & 0.98 & 1.00 & 1.00 & \textcolor{red}{0.02} & 0.99 & 0.99 & 1.00 & 1.00 & 1.00 \\
Llama-3.1-8B-Instruct & 1.00 & 1.00 & 1.00 & 0.88 & 0.98 & 0.98 & 0.98 & 1.00 & 1.00 & 0.80 & 0.98 & 0.99 & 1.00 &  1.00 & 1.00 \\
Yi-1.5-9B-Chat        & 1.00 & 1.00 & 1.00 & 0.42 & 0.71 & 0.90 & 0.90 & 0.96 & 0.95 & 0.94 & 0.98 & 0.96 & 1.00 & 1.00 & 1.00 \\
Gemma-2-9B-it         & 1.00 & 0.95 & 1.00 & 0.97 & 1.00 & 0.96 & 0.96 & 1.00 & 1.00 & 0.89 & 1.00 & 0.99 & 1.00 & 1.00 & 1.00 \\

\hdashline 

Mistral-large     & 0.84 & \textcolor{red}{0.03} & 0.80 & 0.81 & 0.79 & 0.97 & 0.97 & 0.99 & 1.00 & \textcolor{red}{0.08} & 0.99 & 0.98 & 1.00 & 1.00 & \textcolor{red}{0.02} \\
Cohere-Command    & 0.95 & 0.90 & 0.92 & 0.61 & 0.76 & 0.91 & 0.90 & 0.90 & 1.00 & 0.80 & 0.90 & 0.85 & 1.00 & 0.99 & 0.24 \\
Claude-3-Haiku    & 1.00 & 1.00 & 1.00 & 0.98 & 1.00 & 0.98 & 0.98 & 1.00 & 1.00 & 1.00 & 0.99 & 0.99 & 1.00 & 1.00 & 1.00 \\
Claude-3.5-Sonnet & 1.00 & 1.00 & 1.00 & 0.99 & 1.00 & 0.99 & 0.99 & 1.00 & 1.00 & 1.00 & 1.00 & 1.00 & 1.00 & 1.00 & 1.00 \\
GPT-3.5-Turbo     & 1.00 & 0.97 & 1.00 & 0.97 & 0.99 & 0.98 & 0.98 & 1.00 & 1.00 & 0.97 & 1.00 & 1.00 & 1.00 & 1.00 & 1.00 \\
GPT-4o            & 1.00 & 1.00 & 1.00 & 0.99 & 1.00 & 0.99 & 0.99 & 1.00 & 1.00 & 1.00 & 1.00 & 1.00 & 1.00 & 1.00 & 1.00 \\

\bottomrule
           
\end{tabular}
}
\label{tab:parsing}
\caption{Success rate of LLMs generating a value in the correct format for each prediction task. *Origin refers to the country of origin, which is expected to be in ISO-3 format, similarly as the country of nationality. We highlight in red cases were the LLM fails to produce a correctly parsed output in more than 80\% cases.}
\end{table*}